\documentclass[journal,,twocolumn,transmag]{IEEEtran}
\usepackage{caption}
\usepackage{times}
\usepackage{epsfig}
\usepackage{graphicx}
\usepackage{amsmath}
\usepackage{amssymb}
\usepackage{bm}

\ifCLASSINFOpdf
\else
\fi
\begin{document}
%
\title{Local Shrunk Discriminant Analysis (LSDA)}


\author{\IEEEauthorblockN{Zan Gao\IEEEauthorrefmark{1,2},
Guotai Zhang\IEEEauthorrefmark{1,2},
Feiping Nie\IEEEauthorrefmark{3*} and
Hua Zhang\IEEEauthorrefmark{1,2},~\IEEEmembership{Member,~IEEE}}
\IEEEauthorblockA{\IEEEauthorrefmark{1}
Key Laboratory of Computer Vision and System,Ministry of Education,Tianjin University of Technology,Tianjin,300384,China}
\IEEEauthorblockA{\IEEEauthorrefmark{2}Tianjin Key Laboratory of Intelligence Computing and Novel Software Technology,Tianjin University of Technology,300384, China}
\IEEEauthorblockA{\IEEEauthorrefmark{3}Center for OPTical IMagery Analysis and Learning(OPTIMAL), Northwestern Polytechnical University, Xi��an, Shanxi, 710072, China}}

\markboth{Journal of \LaTeX\ Class Files,~Vol.~**, No.~**, May~2017}%
{Shell \MakeLowercase{\textit{et al.}}: Bare Demo of IEEEtran.cls for IEEE Transactions on Magnetics Journals}
%




\maketitle
\begin{abstract}
Abstract:Dimensionality reduction is a crucial step for pattern recognition and data mining tasks to overcome the curse of dimensionality. Principal component analysis (PCA) is a traditional technique for unsupervised dimensionality reduction, which   is often employed to seek a projection to best represent the data in a least-squares sense, but if the original data is nonlinear structure, the performance of PCA will quickly drop. An supervised dimensionality reduction algorithm called Linear discriminant analysis (LDA) seeks for an embedding transformation, which can work well with Gaussian distribution data or single-modal data, but for non-Gaussian distribution data or multimodal data, it gives undesired results. What is worse, the dimension of LDA cannot be more than the number of classes. In order to solve these issues, Local shrunk discriminant analysis (LSDA) is proposed in this work to process the non-Gaussian distribution data or multimodal data, which not only incorporate both the linear and nonlinear structures of original data, but also learn the pattern shrinking to make the data more flexible to fit the manifold structure. Further, LSDA has more strong generalization performance, whose objective function will become local LDA and traditional LDA when different extreme parameters are utilized respectively. What is more, a new efficient optimization algorithm is introduced to solve the non-convex objective function with low computational cost. Compared with other related approaches, such as PCA, LDA and local LDA, the proposed method can derive a subspace which is more suitable for non-Gaussian distribution and real data. Promising experimental results on different kinds of data sets demonstrate the effectiveness of the proposed approach\footnote{This work was supported in part by the National Natural Science Foundation of China (No.61572357, No.61202168). Zan Gao, Guotai Zhang and Hua Zhang is with Key Laboratory of Computer Vision and System,Tianjin University of Technology, Ministry of Education,Tianjin Key Laboratory of Intelligence Computing and Novel Software Technology, Tianjin University of Technology, Tianjin, 300384, China.
Feiping Nie is with the Center for OPTical IMagery Analysis and Learning
(OPTIMAL), Northwestern Polytechnical University, Xi’an, Shanxi, 710072, China. (E-mail: feipingnie@gmail.com)}.
	
\end{abstract}


\maketitle

\IEEEdisplaynontitleabstractindextext

%
\IEEEpeerreviewmaketitle

\section{Introduction}
%
%
%
%
\IEEEPARstart
In pattern recognition and data mining tasks, we are often confront with the curse of dimensionality, which may make it hard for us to train a stable classifier and it will take a long time to train the classifier. Thus, dimensionality reduction is a hot and classical topic [1], which attempts to overcome the curse of the dimensionality and to extract relevant features [2], [3]. For example, although the dimension of original feature of all images of the same subject is very high, its intrinsic dimensionality is usually very low [4]. In dimensionality reduction [5], [6], feature selection [7], [8], [9], [10], where a subset of features of the original set are selected, and feature transformation [11], [12], [13], [14], where the original features are transformed to a new feature subspace, are the two main ways. In contrast to feature selection, feature transformation will obtain much more compact representation of the variables. So far, many dimensionality reduction approaches have been proposed which can be categorized into supervised learning (e.g LDA [11]; Kernel LDA [12]; DiscLDA [13]; LPP [14]; SDRHF [15], Local fisher discriminant analysis (LFDA) [31]) and unsupervised learning (PCA [16], [17]; Kernal PCA [18];  LLE [19]; LPP [14]; NPE [20]; Isomap [22]; Laplacian eigenmaps, [21], [22]) dimensionality reduction methods. The difference between supervised and unsupervised dimensionality reduction algorithms lies in whether the ground truth is utilized or not in learning the transformation matrix [23]. If the ground truth is employed in the subspace learning, the method belongs to the supervised learning method, otherwise, it will be unsupervised learning method. The related references (PCA, [16], [17]; LDA [11]; DiscLDA [12], LFDA [31] show that the supervised learning methods can obtain much better performance than the unsupervised learning methods, and has been applied into different research domains. Among the supervised learning approaches, the most popular and successful one is LDA, which is very suitable for the Gaussian distribution data, and LDA performs well in many applications, but LDA also has some drawbacks: 1) it is only suitable for Gaussian distribution data, since the objective function of LDA is to make the distance between different categories as far as possible, and the same categories as close as possible, thus, when the non-Gaussian distribution data is utilized, the projection direction will be wrong; 2) the dimension of LDA is limited, which must be smaller than the number of classes. If there are only two classes in our task, thus, the dimensionality reduction must be one dimension, which may not represent the original data distribution.

In addition, in our daily life, there are so many unlabeled data on internet, which is very helpful for our future life, thus, unsupervised learning dimensionality reduction algorithms also play an important role. Among these methods, different motivations and objective functions are designed. For example, Turk et al. [16], [17] proposed Principal Component Analysis which is the most frequently used dimensionality reduction method. The motivation of PCA seeks a projection which can best represent the data in a least-squares sense. He et al. [14], [20] proposed Locality Preserving Projections (LPP) in 2003 and Neighborhood Preserving Embedding (NPE) in 2005. Both of them are different from PCA which aims at preserving the global Euclidean structure, but they are linear projective maps whose motivations optimally preserve the neighborhood local structure of the data set. In addition, several nonlinear dimensionality reduction approaches are proposed, such as, locally linear embedding (LLE), isometric feature mapping (Isomap) and Laplacian Eigenmaps, which also preservers the neighborhood relation of data points. However, the motivations between PCA, LPP, NPE and LLE, Isomap and Laplacian Eigenmaps, are very different, and the original LLE, Isomap and Laplcacian Eigenmaps cannot deal with the out-of-sample problem directly [24], that is to say, they only can deal with the training samples, and obtain the low dimension embedding, but for test samples, they cannot directly calculated, analytically or cannot calculated at all. As for PCA, LPP and NPE, they can easily and directly calculate the low dimensional embedding for both training samples and testing samples.

Recently, pattern shrinking [25], [26] is often utilized in clustering algorithms, which not only characterizes the linear and nonlinear structures of data, but also reflects the requirements of clustering, what is more, it can obtain satisfying performance on high dimensional data and non-Gaussian distribution data [30], [32], [33], [34]. Inspired by them, in this paper, we propose a new general unsupervised and supervised learning dimensionality reduction algorithm, called Local shrunk discriminant analysis, where the shrunk pattern and the projection matrix are simultaneously optimized in our objective function, and whose neighborhood structure can be preserved in the dimensionality reduced space. Since the shrunk pattern is utilized in our model, which makes the data more flexible to fit the manifold structure, thus, our proposed model is suitable for non-Gaussian distribution data and the dimensionality reduction is irrelative to the number of classes. What is more, a new efficient optimization algorithm is introduced to solve the non-convex objective function with low computational cost. Promising experimental results on different kinds of datasets demonstrate the effectiveness of the proposed approach. It is worthwhile to highlight the following merits of our work:
\begin{itemize}
\item The proposed algorithm is more capable of uncovering the manifold structure. Particularly, the shrunk pattern does not have the orthogonal constraint, making it more flexible to fit the manifold structure.
\item The pattern shrinking and transformation matrix are simultaneously learnt, which will make the pattern shrinking more suitable for the transformation matrix.
\item The transformation matrix not only learns from the original data points, but also learns from their pattern shrinking, which will make the pattern shrinking more convenient to find a suitable subspace for dimensionality reduction.
\end{itemize}

The rest of this paper is organized as follows. The related work will be given in Section II, and then we detail our proposed algorithm. After that, extensive experimental results are introduced and Section V concludes this paper.

\section{Related Work}
\subsection{Linear Discriminant Analysis}
Since the objective function of LDA is to make the distance between different categories as far as possible, and the same categories as close as possible. Thus, assume there are $n$ samples for dimensionality reduction, and an $r-$dimensional feature vector is utilized to represent each sample, i.e,  $\{x_1,x_2,...,x_n\}$ where $x_i\in R^r$ for $i = 1,2,...n$ . The learning subsapce of $\{x_1,x_2,...,x_n\}$ is represented by $\{z_1,z_2,...,z_n\}$, where  $z_i\in R^d$ for $i = 1,2,...n$  , and $d$ is the dimension of learning subspace. The goal of subspace learning is to find a optimized transformation matrix $W\in R^{r\times d} $, and each sample is then projected into a low-dimensional subspace by $z_i = W^Tx_i$. Denote $X =\{x_1,x_2,...,x_n\}$  and $Z =\{z_1,z_2,...,z_n\}$  , therefore, $Z = W^TX$ . As for $n$ sample, we assume that each image belongs to one of $C$ classes, thus, these original data can also be represented by $\{X_1,X_2,...,X_c\}$, where the number of each class $X_i$ is $n_i$ . And then, the mean value $\mu$ of all samples are computed, after that, the mean value $\mu_i$ of each class $X_i$ be also calculated, thus, let the between-class scatter matrix be defined as

   \begin{equation}
    S_{B}= \sum_{i = 1}^C n_{i}(\mu_{i}-\mu)(\mu_{i}-\mu)^{T}
   \end{equation}

And the within-class scatter matrix be defined as
    \begin{equation}
      S_{W}=\sum_{i=1}^C\sum_{x_{k}\in{X_{i}}} (x_{k}-\mu_{i})(x_{k}-\mu_{i})^T
    \end{equation}
Where $X_i$ denotes the set of samples in class $i$. The goal of LDA is also to obtain a transformation matrix $W\in R^{r\times d}$, and the low-dimensional feature vector can be calculated by $z_i = W^Tx_i$. If $S_W$ is nonsingular, the optimal projection $W_{opt}$ is chosen as the matrix with orthonormal columns which maximizes the ratio of the determinant of the between-class scatter matrix of  the projected samples to the determinant of the within-class scatter matrix of the projected samples, i.e.,
   \begin{equation}
     W_{opt} = \arg\max_W\\\frac{|W^TS_BW|}{|W^TS_WW|} = [w_1,w_2,...w_m]
   \end{equation}
Where $\{w_i|i = 1,2,...,m\}$ is the set of the generalized eigen-vectors of $S_B$ and $S_W$ corresponding to the $m$  largest generalized eigen-values $\{\lambda_i|i = 1,2,...,m\}$  , i.e.,
    \begin{equation}
     S_Bw_i = \lambda_i S_Ww_i, i = 1,2,...,m
    \end{equation}
Note that there are at most $C-1$ nonzero generalized eigen-values, and so an upper bound on $m$ is $C-1$, where $C$ is the number of classes.

\subsection{Locality Preserving Projections}
The linear projective maps (LPP), which is a variational problem, optimally preserves the neighborhood structure of the data set. In fact, LPP also can be seen as an alternative to PCA, which is a classical linear technique that projects the data along the directions of maximal variance. Thus, the objective function of LPP is defined as follows:
                \begin{equation}
                   W_{opt}=\arg\min_{W^TXDX^TW = I}\\ tr(W^TXLX^TW)
                \end{equation}
Where $a_{ij}$ is utilized to measure the similarity of $x_i$ and $x_j$. The similarity matrix [28], [29] $A\in R^{n\times n}$  is composed of all the $a_{ij}$, which is utilized to characterize the manifold structure of original data points, i.e. ${x_1,x_2,...,x_n}$ . In detail, a $k$-nearest neighborhood graph is firstly constructed, and the points are considered as nodes, then connecting every point to its k nearest neighbors is utilized as theirs edges.

According to whether the label information is utilized or not, the construction of $a_{ij}$ is divided into supervised learning and unsupervised learning. In supervised learning, denote that $N^s(x_i)$ is the index set of the points, which are $k$ nearest neighbors of $x_i$, and the class of these points are same to $x_i$. Thus, the weight matrix $A$ in supervised learning, associated with this $k$ -nearest neighborhood graph, is computed by the following equation:
\begin{equation}
\left.
\!\!\!\!\!\!a_{ij} \!= \!\!a_{ji}\! =\!\!
\bigg\{
\begin{split}
&e^{-\frac{\parallel x_i - x_j \parallel_{2}^2}{2\sigma^2}}
 \qquad x_i\in N^s(x_j) \ or \  x_j \in N^s(x_i)\\
&0 \quad \ otherwise
\end{split}
\quad
\right.
\end{equation}
Where $\sigma$ is the width parameter to control the Gaussian distribution.

As for unsupervised learning, denote that $N^u(x_i)$ is the index set of the points, which are $k$ nearest neighbors of $x_i$ and the classes of these points can be from different classes. Thus, the weight matrix $A$ in unsupervised learning can be defined as follows:
\begin{equation}
\left.
\!\!\!\!\!\!a_{ij} \!= \!\!a_{ji}\! =\!\!
\bigg\{
\begin{split}
&e^{-\frac{\parallel x_i - x_j \parallel_{2}^2}{2\sigma^2}}
 \quad x_i\in N^u(x_j) \ or \  x_j \in N^u(x_i)\\
&0 \qquad \ otherwise
\end{split}
\qquad
\right.
\end{equation}

In addition,$L = D-A$ , where $D$ is a diagonal matrix with entries $D_{ij} = \sum_j A_{ij}$ . In fact, it is the Laplacian matrix of the above defined $k$-nearest graph with weight matrix S. Specially, in the construction of similarity matrix, if $k$-nearest neighborhood does not contain the label information of samples, thus, LPP is considered as unsupervised learning dimensionality reduction, or it will be known as supervised learning dimensionality reduction.

For the optimization of Eq. (5), it can be obtained by computing the eigenvectors and eigenvalues for the generalized eigenvector problem:
   \begin{equation}
               XLX^TW = \lambda XDX^TW
               \end{equation}
Let the columns vectors $w_0,w_1,...,w_{d-1}$ be the solutions of Eq. (5), which are ordered by their eigenvalues, $\lambda_0<\lambda_1<...<\lambda_{d-1}$. Thus, the dimensionality reduction can be obtained by
   \begin{equation}
          x_i\to y_i = W^Tx_i, W = (w_0,w_1,...,w_{d-1})
               \end{equation}

\subsection{Subspace Learning via Pattern Shrinking}	
Assume there are $n$ samples for clustering, and an $r-$dimensional feature vector is utilized to represent each sample, i.e,$\{x_1,x_2,...,x_n\}$ where $x_i\in R^r$ for $i = 1,2,...n$. The shrunk pattern of these samples are defined by $\{y_1,y_2,...,y_n\}$ , where $y_i\in R^r$ for $i = 1,2,...n$. The learning subspace of $\{x_1,x_2,...,x_n\}$ is represented by $\{z_1,z_2,...,z_n\}$, where $z_i\in R^d$ for  $i = 1,2...n$, and $d$ is the dimension of learning subspace. The goal of subspace learning is to find a optimized transformation matrix $W\in R^{r\times d}$, and each sample is then projected into a low-dimensional subspace by $z_i =W^Tx_i$. Denote $X = \{x_1,x_2,...,x_n\}$  $Y = \{y_1,y_2,...,x_n\}$ and  $Z = \{z_1,z_2,...,z_n\}$ , therefore, $Z = W^TX$ or $Z = W^TY$ . Since the manifold structures of the original data may be nonlinear, but the linear projection approaches cannot fully consider about them. Thus, the shrunk patterns are employed in the model, which can make the data more flexible to fit the manifold structure. The model is defined as follows:
\begin{equation}
\begin{aligned}
\arg&\min_{ W^TW = I,Z = W^TY}\! \!\!  \{ (1-\delta) \sum_{i = 1}^n\sum_{j = 1}^na_{ij} \parallel \! \! y_i - y_j \! \! \parallel_2^2\\
&+\delta\sum_{i = 1}^n\! \! \parallel x_i - y_i\! \! \parallel^2_2 - \beta\sum_{i = 1}^n\! \! \parallel z_i - \frac{1}{n}\sum_{j=1}^nz_j\! \! \parallel^2_2\}
\end{aligned}
\end{equation}
Where $\delta$ and $\beta$ are used to control the balance of each regular term. The meaning of other parameters in Eq.(10) is same to Eq.(5) and Eq.(6).

In this objective function, there are three kinds of regularization terms, and the reason why the first term is utilized, is that nearby points are more likely to belong to the same cluster, and the shrunk pattern should maintain the similarities of the original data, which is measured by a weight matrix; As for the second regular term, the reason is that the consistency between the original data and their shrunk pattern should be kept; Finally, after learning the shrunk pattern, a subspace is expected to learn, where the projections of shrunk patterns should maintain their original separations. It guarantees that the information loss in deriving shrunk data is as little as possible.


\section{LOCAL SHRUNK DISCRIMINANT ANALYSIS}
In this section, we will introduce some notations and formulate the dimensionality reduction approach by our pattern shrinking technique. After that, we show how to derive the approximated solution in a quick way. Finally, some preliminary discussions are provided.
\subsection{Problem Formulation}
In our real-life, there are often non-Gaussian distribution data, when LDA is employed to reduce the dimension, the projection direction may be wrong. Fortunately the pattern shrinking  [25], [26] may be helpful for us to find a suitable transformation matrix, thus, the pattern shrinking is employed in our model. In addition, in subspace learning via pattern shrinking, it cannot directly deal with the out-of-the-sample problem [24], where only the low dimensional embedding map of training samples can be calculated but the samples out of the training set (i.e. testing samples) cannot be computed directly, thus, the transformation matrix is simultaneously optimized from the original features and shrunk pattern. For the construction of this objective function, our motivations are as follows:
\begin{itemize}
\item[1)]  Basic assumption of clustering or subspace learning. Intuitively, if two points are nearby, they should belong to the same cluster and the similarity weight should be large. On the contrary, if two points are far away, the corresponding similarity weight should be small. This objective can be implemented by employing the similarity matrix $S$. After obtaining the pattern shrinking, the similarity between the shrunk patterns $y_i$ and $y_j$ should keep the consistency with the similarity between the original data $x_i$  and $x_j$.
\item[2)] The shrunk pattern should keep the consistency with the original data. More concretely, the shrunk pattern and the original data should not be far away. Compared with the first motivation, which uses the local similarity for shrinking data, this objective function can be regarded as keeping dissimilarity in firtst motivation, and we also require that the pattern shrinking data should be close to the original data.
\item[3)] After learning the shrunk pattern, we expect to learn a projection matrix in which the similarity of the projections between different shrunk patterns, and the difference of the projections between the original data points and shrunk pattern, should maintain their original separations. By this way, it guarantees that the loss of information in deriving shrunk data and the embedding is as little as possible. In addition, the projection matrix and the pattern shrinking are simultaneously optimized, which makes the project matrix more suitable for dimensionality reduction.
\end{itemize}
For the simplification of writing, the notation in Section II is utilized. In order to keep the local similarity, the shrunk pattern should inherit the local similarity of the original data. Moreover, the shrunk pattern should not be far way from the original data. Thus, it also requires that the shrunk pattern consists with the original data. Thus, the following loss function can be directly minimized:

\begin{equation}
\begin{aligned}
\arg\min_{Z}\! \!\!  &\sum_{i = 1}^n\sum_{j = 1}^n \! \!a_{ij}\! \! \parallel \! \! z_i - z_j \! \! \parallel_2^2
&+\gamma\sum_{i = 1}^n\! \! \parallel x_i - z_i \! \! \parallel^2_2
\end{aligned}
\end{equation}

Although the shrunk pattern can well represent the original data, the dimension may be very high, and it will be difficlut to train classifiers or other computations. Thus,we hope the learnt feature representation not only can has robust feature representation, but also it can has low dimension. Thus, the loss function can be defined as follows:

\begin{equation}
\begin{aligned}
\arg\min_{ W^TS_tW = I,Z}\! \!\!  &\sum_{i = 1}^n\sum_{j = 1}^n \! \!a_{ij}\! \! \parallel \! \! W^Tz_i - W^Tz_j \! \! \parallel_2^2\\
&+\gamma\sum_{i = 1}^n\! \! \parallel W^Tx_i - W^Tz_i \! \! \parallel^2_2
\end{aligned}
\end{equation}

Where $S_t\in R^{r\times r}$ is the covariance matrix of the original data points.  $I\in R^{d\times d}$ is the unit matrix, and $\gamma$ is used to control the balance between the original data and shrunk patterns. As for other parameters, their means are same to Eq. (10). In this objective function, the first term is employed to meet the first motivation, and then the second motivation is satisfied by the second term. As for last motivation, it is achieved by $W$ in each regularization term.

\subsection{Optimization}
Since the Eq.(12) is non-convex, thus, it is difficult for us to directly optimize it. In order to simply write, we denote that $f_i = W^Tz_i \in R^d$ and $f_i = W^Tz_i \in R^d$ are the embedding of corresponding $z_i$ and $z_j$, thus, the problem of Eq.(12) is equivalent to
\begin{align}
\arg\min_{\! \! \! \! \! \! \!  W^TS_tW = I,F}\! \!\! \sum_{i = 1}^n\sum_{j = 1}^n \! \! a_{ij}\! \! \parallel \! \! f_i - f_j \! \! \parallel_2^2+\gamma\sum_{i = 1}^n\! \! \parallel W^Tx_i - f_i\! \! \parallel^2_2
\end{align}
Where $F = \{f_1,f_2...f_i...f_n\}\in R^{d\times n}$ is the embedding matrix of corresponding $\{z_1,z_2...z_n\}$. However, the problem is still non-convex, it seems difficult to find the optimal solution, even to find a solution, since:

When fix $W$, the original problem equivalent to :
\begin{align}
\arg\min_{F}\! \!\!  \sum_{i = 1}^n\sum_{j = 1}^n \! \!a_{ij}\! \! \parallel \! \! f_i - f_j \! \! \parallel_2^2+\gamma\sum_{i = 1}^n\! \! \parallel W^Tx_i - f_i\! \! \parallel^2_2
\end{align}
When fix $F$, the new objective function will be:
\begin{align}
\arg\min_{ W^TS_tW = I}\! \sum_{i = 1}^n \! \!  \parallel \! \! W^Tx_i - f_i \! \! \parallel_2^2
\end{align}
However, for the optimization of Eq.(15), it is still difficult for us to obtain the optimization solution. Interestingly, we can find the closed form (and thus optimal) solution to the problem of Eq.(13).
First, the objective function in Eq.(13), becomes
\begin{align}
\arg\min_{ W^TS_tW = I,F}Tr(F^TLF)+\gamma\parallel X^TW- F\parallel^2_F
\end{align}
Where $L$ is the so called graph Laplacian induced from the graph structure. Specifically, $L = D-A$ , where $A$ is the pre-computed similarity matrix among the original data, and $D$ is a diagonal matrix, whose element can be obtained by $D_{i,i} = \sum_iA_{i,j}$ .As for $\parallel\bullet\parallel_F$ , it denotes the $F-norm$.
Denote $\Gamma(W,F)=Tr (F^TLF)+\gamma\parallel X^TW- F\parallel^2_F$ , and setting the gradients with respect to $F$  to zero, and we can calculate
\begin{align}
\frac{\partial\Gamma (W,F)}{\partial F}=0\Rightarrow F=\gamma (L+\gamma I)^{-1}X^TW
\end{align}
Substituting $F=\gamma (L+\gamma I)^{-1}X^TW$ into the Eq.(16), we can obtain
\begin{align}
\arg\min_{W^TS_tW = I}\!Tr(W^TX(\gamma I-\gamma^2 (L+\gamma I)^{-1})X^TW)
\end{align}
For the optimization of Eq.(18), the constrained minimization can then be done using the method of Lagrange multipliers:
\begin{equation}
\begin{aligned}
\Gamma(W) = &Tr(W^TX(\gamma I-\gamma^2(L+\gamma I)^{-1})X^TW)\\
&+\lambda (I-W^TS_tW)
\end{aligned}
\end{equation}
Setting the gradients with respect to $W$ to zero, we have

\begin{equation}
\begin{aligned}
\frac{\partial\Gamma(W)}{\partial W}=&2X(\gamma I-\gamma^2(L+\gamma I)^{-1})X^TW)\\
&-2\lambda S_tW)=0
\end{aligned}
\end{equation}
By defining
\begin{align}
H=X(\gamma I-\gamma^2(L+\gamma I)^{-1})X^T
\end{align}
The transformation vector in Eq.(18) that minimizes the objective function is given by the minimum eigenvalue solution to the following generalized eigenvector problem
\begin{align}
HW = \lambda S_tW
\end{align}
Since $S_t$ is the covariance matrix of the original data points, thus, it may be nonsingular, thus, the Generalized Singular Value Decomposition (GSVD) is employed.
Note that  at most $C-1$ nonzero generalized eigen-valuse in LDA is not requested in our model. After obtaining the projection vector,$X_i$  can be mapped to a low dimensional space $z_i$ by $z_i = W^Tx_i$ .

\subsection{Algorithm Analysis}
In LSDA model, it not only can make the data more flexible to fit the manifold structure, which will be useful for non-Gaussian distribution data, but also it is the generalized form of LDA.  In what follows, we will introduce the form of objetive function when different parameters are utilized, such as,  $\gamma \to \infty $, $\gamma\to0$ and $\gamma\in R$, and we will carefully explain it in the following section.

When \bm{$\gamma\to\infty$} , thus, the second regular term in Eq.(16) will be zero, and the optimization of Eq.(16) is equal to
\begin{equation}
\begin{aligned}
\arg\min_{W^TS_tW = I,F}& Tr(F^TLF)\\
 s.t. F=X^TW
\end{aligned}
\end{equation}
Substituting $F=X^TW$ into the Eq.(23), we can obtain
\begin{align}
\arg\min_{W^TS_tW = I}Tr(W^TXLX^TW)
\end{align}
In fact, this objective function is one kind of local LDA.

When \bm{$\gamma\to0$}, it is clear that the optimization solution is $Tr(F^TLF) = 0$, and then we just optimize
\begin{align}
\arg\min_{W^TS_tW = I,F}\sum_{i = 1}^n\parallel W^Tx_i - f_i\parallel^2_2
\end{align}
Assume there are $P$ connected component in matrix $L$, in which it is obvious that $P$ is much bigger than the number of classes $C$, and for class $k$, the number of its connected component is $v_k$, thus, the connected components of class $k$ can be denoted by $\{C_k^1,C_k^2,...,C_k^{v_k}\}$. Among all connected component, $f_i = f_j$ . If there is $x_i\in C^l_k$ , thus $f_i = f_k^l$. Therefore, the optimization of Eq.(23) becomes
\begin{align}
\arg\min_{W^TS_tW = I,F}\sum_{k = 1}^c\sum_{l = 1}^{v_k}\sum_{x_i\in C_k^l}\parallel W^Tx_i - f_k^l\parallel^2_2
\end{align}
We can obviously know that the optimization solution is $f_k^l = \frac{1}{|C_k^l|}\sum_{x_i\in C^l_k}W^Tx_i$ . If we assume $m_k^l = \frac{1}{|C_k^l|}\sum_{x_i\in C^l_k}x_i$, thus, the objective funcition Eq.(26) can be represented by
\begin{align}
\arg\min_{W^TS_tW = I,F}\sum_{k = 1}^c\sum_{l = 1}^{v_k}\sum_{x_i\in C_k^l}\parallel W^Tx_i - W^Tm_k^l\parallel^2_2
\end{align}
In fact, from the objective function, we can know that it is another local LDA [31]. Further, when $v_k = 1$ , the objective function will become
\begin{align}
\arg\min_{W^TS_tW = I,F}Tr(W^TS_wW)
\end{align}
It can be observed that it is the traditional LDA.

Furthermore, for any \bm{$\gamma$}, when $v_k = 1$ and $a_{ij}$ is required to

\begin{equation}
\left.
\!\!\!\!\!\!a_{ij} \!=
\bigg\{
\begin{split}
&\frac{1}{n_k} \qquad  x_i,x_j\in C_k \\
&0 \qquad \ otherwise
\end{split}
\qquad
\right.
\end{equation}

Since it is obvious that $I-\gamma(L+\gamma I)^{-1}$ is block diagonal matrix, and for each diagonal block, we can know that
\begin{align}
(L+\gamma I)^{-1} = ((1+\gamma)I-\frac{1}{n_i}11^T)^{-1} \\
\nonumber =\frac{1}{1+\gamma}I+\frac{1}{n_i\gamma(1+\gamma)}11^T
\end{align}
\begin{align}
I-\gamma(L+\gamma I)^{-1} = \frac{1}{1+\gamma}(I-\frac{1}{n_i}11^T)
\end{align}

Thus, Eq.(21) will become $X(\gamma I-(\gamma)^2(L+\gamma I)^{-1}X^T = \frac{\gamma}{1+\gamma}S_w$. Thus, as for any \bm{$\gamma$}, the objective function of our proposed method is equal to the traditional LDA.

Thus, we can conclude that LSDA has more strong generalization performance, whose objective function with different extreme parameters will become local LDA and traditional LDA.

\section{EXPERIMENTS AND DISCUSSION}
In order to evaluate the performance of our proposed LSDA, we perform extensive experiments on three different kinds of tasks, such as handwritten digit recognition task, face recognition task and object recognition task. Specially, on face recognition, four face recognition databases are utilized. At the same time, in order to fair comparison, we divide the existing dimensionality reduction algorithms into unsupervised learning (including PCA, NPE, LPP and LSDA with unsupervised learning) and supervised learning algorithms (LDA, LPP,LFDA and LSDA with supervised learning). For the code of PCA, NPE, LDA and LPP, we download them from the internet\footnote{http://www.ews.uiuc.edu/~dengcai2/Data/data.html} , and also strictly keep the parameters same with them.
\subsection{Dataset}
\begin{itemize}
\item USPS dataset\footnote{http://www-i6.informatik.rwth-aachen.de/~keysers/usps.html}

It is a handwritten digit database. A popular subset contains 9298 handwritten digit images in total is employed, where it contains 4649 training images and 4649 test images with 16x16 handwritten digit images, and the number of each class is over 400 samples in the training and testing dataset. Fig.1 shows different handwritten digits.
\begin{figure}[t]
\begin{center}
\includegraphics[width=2in,height = 0.7in]{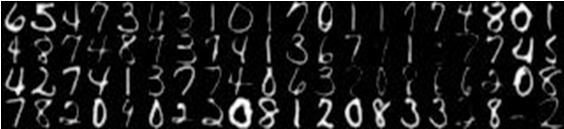}
\caption{Sample images from USPS database}
\end{center}
\end{figure}
\item Yale database (Yale)\footnote{http://cvc.yale.edu/projects/yalefaces/yalefaces.html}

It contains 165 gray scale images of 15 individuals. The images demonstrate variations in lighting condition, facial expression (normal, happy, sad, sleepy, surprised, and wink). Fig.2 shows the 11 images of one individual in Yale data base.
\begin{figure}[t]
\begin{center}
\includegraphics[width=2in,height = 0.7in]{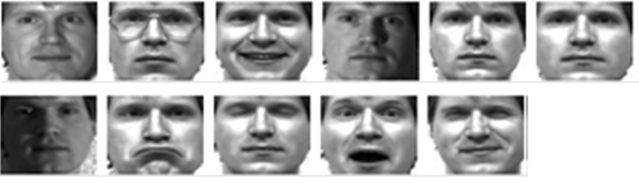}
\caption{ Sample images from Yale database}
\end{center}
\end{figure}
\item  ORL (Olivetti Research Laboratory) database (ORL)\footnote{http://www.uk.research.att.com/facedatabase.html}

It contains 400 images of 40 individuals. Some images were captured at different times and have different variations including expression (open or closed eyes, smiling or non-smiling) and facial details (glasses or no glasses). The images were taken with a tolerance for some tilting and rotation of the face up to 20 degrees. 10 sample images of one individual in the ORL database are displayed in Fig.3.
\begin{figure}[t]
\begin{center}
\includegraphics[width=2in,height = 0.7in]{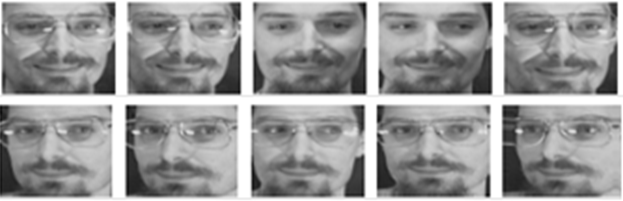}
\caption{Sample images from ORL database}
\end{center}
\end{figure}

\begin{figure}[t]
\begin{center}
\includegraphics[width=2in,height = 0.7in]{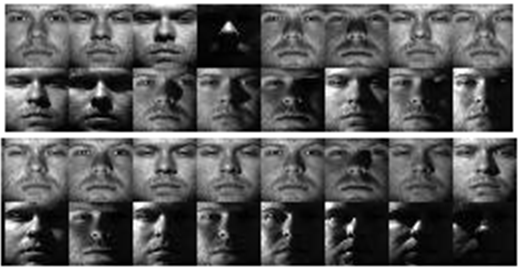}
\caption{Sample images from YaleB database.}
\end{center}
\end{figure}
\item  Extended Yale Face Database B (YaleB)\footnote{http://vision.ucsd.edu/~leekc/ExtYaleDatabase/ExtYaleB.html}

The original YaleB database contains 16128 images of different human subjects under 9 poses and 64 illumination conditions, but in our experiments, 38 individuals around 64 near frontal images under different illuminations per individual are utilized, whose sample number is 2414, and the images are showed in Fig.4
\item  CMU-PIE\footnote{http://www.ri.cmu.edu/research\_project\_detail.html?project\_id=418
\&menu\_id=261}

The CMU PIE face database contains 68 individuals with 41,368 face images as a whole. The face images were captured by 13 synchronized cameras and 21 flashes, under varying pose, illumination and expression. We choose the five near frontal poses (C05, C07, C09, C27, C29) and use all the images under different illuminations, lighting and expressions which leaves us 170 near frontal face images for each individual. Fig.5 shows their samples.

\begin{figure}[t]
\begin{center}
\includegraphics[width=2in,height = 0.7in]{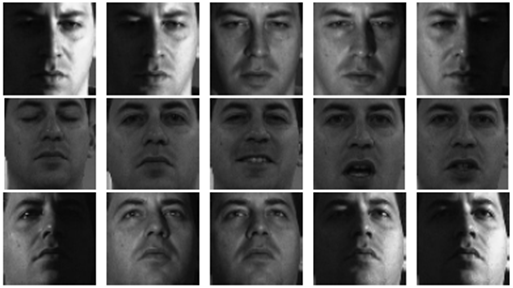}
\caption{Sample images from CMU-PIE database}
\end{center}
\end{figure}

\item  Coil-100\footnote{http://www.cs.columbia.edu/CAVE/software/softlib/coil-100.php}

It contains 100 objects. The images of each object were taken 5 degrees apart as the object is rotated on a turntable and each object has 72 images.
\begin{figure}[htbp]
\begin{center}
\includegraphics[width=2in,height = 0.7in]{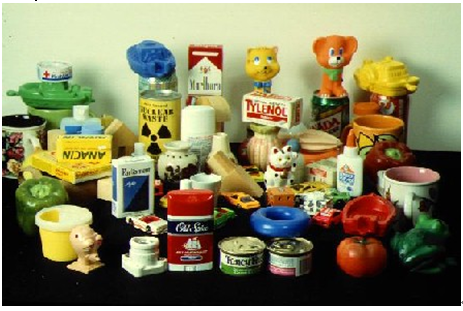}
\caption{Sample classes from Coil-100 database}
\end{center}
\end{figure}
\end{itemize}

\subsection{Pre-processing Step}
All images are manually aligned and cropped, and expect for USPS dataset, the size of each cropped image is 32x32 pixels, with 256 gray levels per pixel. The pixel values are then scaled to [0,1] (divided by 256). For the vector-based approaches, the image is represented as a 1024-dimesional vector. For classification, the nearest-neighbor classifier for its simplicity is utilized in our experiments, and following most work on recognition task, we adopt recognition accuracy as our evaluation metrics in our experiments. In the computation of the similarity matrix, the Euclidean metric is employed as our distance measure.

\subsection{Parameter Setup}
For unsupervised learning, the size of neighborhood k is set to 30 for all the dimensionality reduction algorithms, but for supervised learning, the size of neighborhood k is decided by the number of samples of each subject, and if the number of samples of one subject is small than 50, thus, k is set to its number of samples, otherwise, k is equal to 50.

In all experiments, we tune the sigma parameter in the range of [0.1, 0.2, 0.3, 0.4, 0.5, 0.6, 0.7, 0.8, 0.9, 1.0], and gamma parameter in the range of [$2^{-10}, 2^{-9}, ..., 2^9, 2^{10}$]. Since the accuracies of PCA, NPE and LDA are only affected by the dimension without tuning other parameters, thus, their accuracies are stable for each dimension. However, for LPP, its accuracy will be different with the change of sigma parameter, thus, the sigma parameter will be traversed for each dimension, and then the best result is reported. As for LSDA, the change of sigma and gamma parameters will affect the performance of LSDA, thus, the alternating method is utilized [27], where at each iteration, we first fix the one variable, and then optimize the other variables, at the same time, we update these variables by repeated iterative, and then keep the best accuracy.

Note that the results of recognition algorithms vary on dataset split, in order to reduce the influence of statistical errors, different training and testing datasets are constructed, where different number of images are chosen for different datasets according by the number of samples. In our experiments, we randomly choose L (=2, 3, 4, 5, 6, 7, 8) images for Yale and ORL databases, L (=5, 10, 20, 30, 40, 50) images for YaleB and Coil-100 databases, and L (=5, 10, 20, 30) images for CMU-PIE database per individual to form the training set, and then the rest of them is utilized to form the testing dataset. What is more, for each given L, we will repeat 50 times, and then average the results over 50 random splits. As for USPS dataset, the fixed training and testing datasets are given, thus, we also strictly follow it.
\subsection{Toy Example}
Taking Fig. 7 as an intuitive example, and from it, we can know that the original data is non-Gaussian data, where there are two class man-made data points, and the number of each ellipse is 1000. Since the goal of LDA is to make the distance between different categories as far as possible, and the same categories as close as possible, thus, when LDA is utilized, the projector direction is wrong. However, in our LSDA model, the shrunk pattern is learnt, which will make the data more flexible to fit the manifold structure, and the project direct is correct.

\begin{figure*}[htbp]
\begin{minipage}[t]{0.33\linewidth}
\centering
\includegraphics[width=2.5in,height = 1.5in]{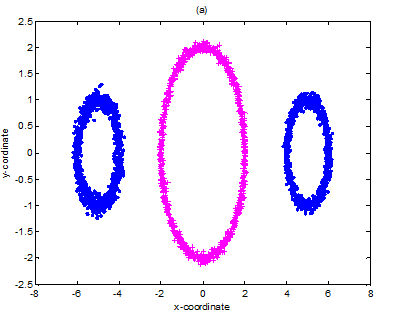}
\end{minipage}%
\begin{minipage}[t]{0.33\linewidth}
\centering
\includegraphics[width=2.4in,height = 1.5in]{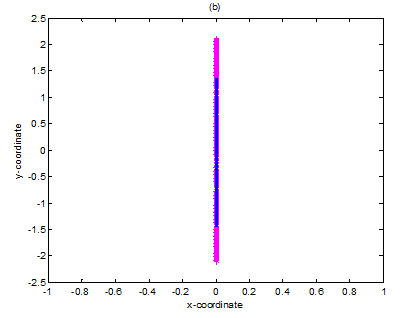}
\end{minipage}
\begin{minipage}[t]{0.3\linewidth}
\centering
\includegraphics[width=2.4in,height = 1.5in]{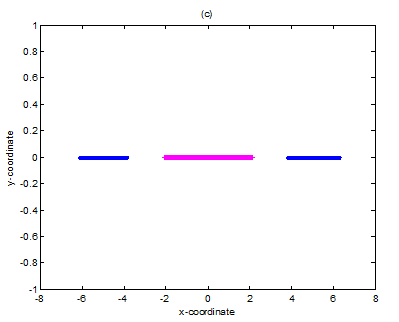}
\end{minipage}

\caption{Toy Examples of dimensionality reduction. (a) the original data is a non-guassian distribution data, where red ellipse data belongs to class one, and other two ellipse datum belong to class two. The mean of all ellipses is zero, but the variances of from left to right in Fig.7(a) are 0.1, 0.05 and 0.07 respectively. The dimension of original data is two, and the dimension of dimensionality reduction data is one; (b) The results of dimensionality reduction by LDA; (c) The results of dimensionality reduction by LSDA where sigma and gamma are set to 0.5 and $2^{-5}$ respectively.}
\vspace{-1em}
\end{figure*}

\subsection{Experimental Results of Unsupervised Learning Methods }
In the construction of similarity matrix $A$, $a_{ij}$ is built by Eq.(7) where the k-nearest neighbor samples without their label information are employed, thus, LSDA is considered as unsupervised learning dimensionality reduction method. Therefore, we will firstly assess the performance of unsupervised learning dimensionality reduction method.  In our experiments, for all algorithms, we will evaluate the performance when the dimensionality changes from 10 to 100 on top four databases, and from 10 to 200 on the last three databases. In addition, in order to fairly and conveniently comparison, if the dimension of these dimensionality reduction algorithms cannot reach 100 or 200, but we will still keep the best performance for the later dimensions, for example, if the maximum dimension of PCA only can obtain 60 whose performance is 85\%, thus, all the performances for 70, 80, 90 and 100 dimensions will be 85\%.  At the same time, for each dataset split, we will repeat 50 times, and then average recognition accuracy is utilized as the evaluation criterion. As for the baseline method in all datasets, the recognition is simply performed in the original 1024-dimensional image space without any dimensionality reduction.
The experimental results of USPS, Yale, ORL, YaleB, CMU-PIE and Coil-100 are showed on Fig.8, Fig.9, Fig.10, Fig.11, Fig.12 and Fig.13 respectively, but since the limitation of the space, only the results of the first and last dataset splits are given for USPS, Yale, ORL, YALE, YaleB, CMU-PIE and Coil-100. In these figures, the horizontal axis is the dimension index, and the vertical coordinates means the average recognition accuracy. At the same time, we also choose the best results from them to compare to baseline method, whose results are given in Table I, Table II, Table III, Table IV, and Table V respectively. In these Tables, the first row denotes different dataset splits, and the most left column means different dimensionality reduction algorithms. As for the data in these tables, they are average recognition accuracy and the standard deviation of them. From these figures and tables, we can observe that:
\begin{itemize}
\item For PCA methods, we can observe that although the performances of PCA on all datasets expect for COIL-100 are lower than the baseline, their performances are relative stable with the variation of different databases, and their accuracies always are comparable with the baseline even when their dimensions are under 100. In fact, with the addition of dimensions, their performance can relatively improve, but the dimension will be very high which will take a long time for recognition task.
\item For NPE reduction method, it can obtain relative good accuracy on Yale, YaleB and ORL databases whose performances are comparable with the baseline, but on USPS, CMU-PIE and COIL-100 database, its performance quickly decreases, which is much worse than the baseline. Thus, the method is not stable.  Furthermore, the performance of NPE is little worse than PCA, but the improve speed of its accuracy is very quick with the increase of the dimension.
\item For LPP with unsupervised learning method, it preserves the neighborhood structure of the data point, and it can obtain good accuracy on USPS, YaleB and CMU-PIE database whose performance is comparable with the baseline, but on Yale, ORL and COIL-100 databases, their performance quickly decreases, which is also much worse than the baseline. The reason why the performance of LPP with unsupervised learning method changes so much, is that different databases have different data distributions, and in some databases, there are some noise points in the neighborhood structure of the data point, and the noise points affect the following recognition  task.
\item For LSDA with unsupervised learning method, although the neighborhood structure of the data points is also preserved in LPP and NPE, the pattern shrinking is employed in LSDA, what is more, the pattern shrinking and the transformation matrix are simultaneously learnt. Experimental results show that with the variation of different databases and the number of dimensions, our proposed LSDA method are almost always consistently the best algorithm, whose improvements can reach about 3\% to 10\% improvement on all databases when comparing with other dimensionality reduction algorithms. Further, when comparing with the baseline on all databases, whose improvement achieves 10\% to 30\%. The experimental results show that our method is very efficient and effective.  In other words, our method is stable and efficient.
\item From these experimental results, we also can know that no matter what kinds of dimensionality reduction algorithms, the number of samples of each subject will affect their performances, and with the increase of the number of samples, their accuracies greatly improve, but for each dataset split, when the dimension is added, their accuracies will keep stable.
\begin{figure}[htbp]
\begin{center}
\includegraphics[width=2in,height = 1.3in]{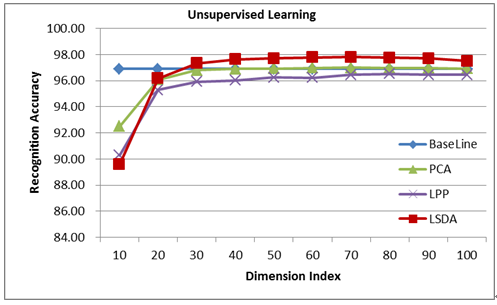}
\caption{Performance comparisons on USPS database with different dimensions. Since the best performance of NPE only is 36\%, which will reduce the difference of other algorithms, thus, its curve is ignored.}
\end{center}
\end{figure}
\end{itemize}

\begin{figure}[htbp]
\begin{minipage}[t]{0.5\linewidth}
\centering
\includegraphics[width=1.7in,height = 1.2in]{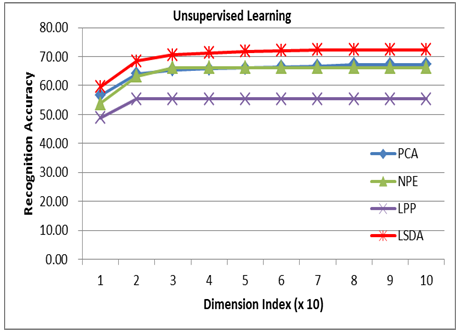}
\end{minipage}%
\begin{minipage}[t]{0.5\linewidth}
\centering
\includegraphics[width=1.7in,height = 1.2in]{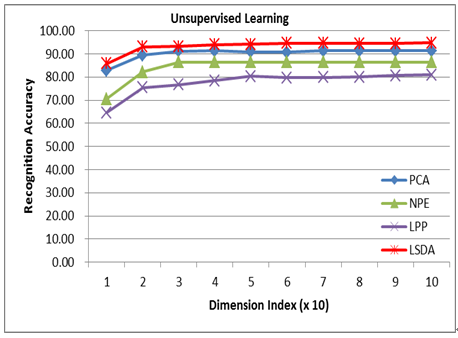}
\end{minipage}
\caption{Performance comparisons on Yale database with different dimensions, and from left to right, they are from 2train and 8train respectively.}
\vspace{-1em}
\end{figure}

\begin{figure}[htbp]
\begin{minipage}[t]{0.5\linewidth}
\centering
\includegraphics[width=1.7in,height = 1.2in]{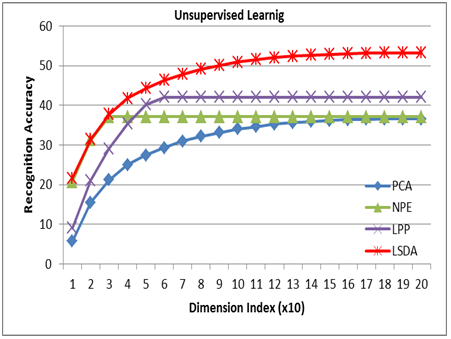}
\end{minipage}%
\begin{minipage}[t]{0.5\linewidth}
\centering
\includegraphics[width=1.7in,height = 1.2in]{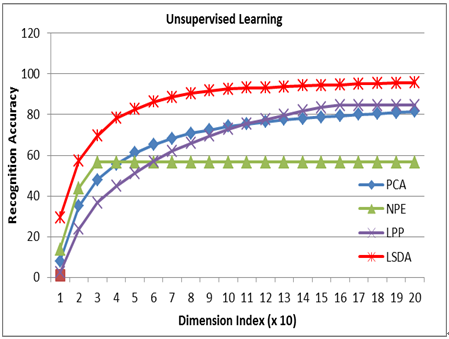}
\end{minipage}
\caption{Performance comparisons on ORL database with different dimensions, and from left to right, they are from 2train and 8train respectively.}
\vspace{-1em}
\end{figure}

\begin{figure}[htbp]
\begin{minipage}[t]{0.5\linewidth}
\centering
\includegraphics[width=1.7in,height = 1.2in]{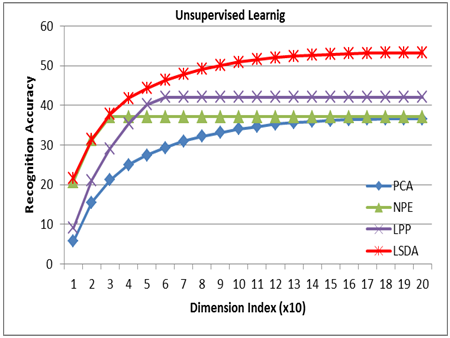}
\end{minipage}%
\begin{minipage}[t]{0.5\linewidth}
\centering
\includegraphics[width=1.7in,height = 1.2in]{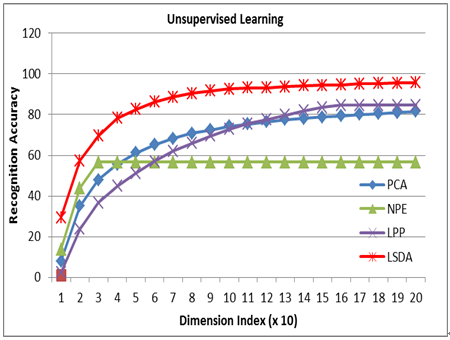}
\end{minipage}
\caption{Performance comparisons on YaleB database with different dimensions, and from left to right, they are from 5train and 50train respectively.}
\vspace{-1em}
\end{figure}

\begin{figure}[htbp]
\begin{minipage}[t]{0.5\linewidth}
\centering
\includegraphics[width=1.7in,height = 1.2in]{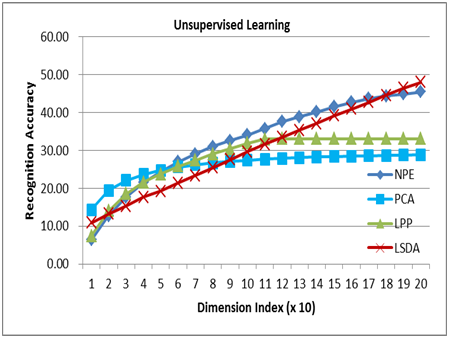}
\end{minipage}%
\begin{minipage}[t]{0.5\linewidth}
\centering
\includegraphics[width=1.7in,height = 1.2in]{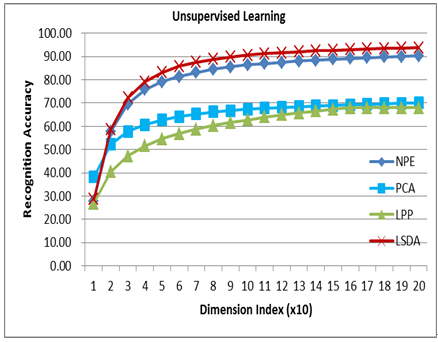}
\end{minipage}
\caption{Performance comparisons on CMU-PIE database with different dimensions, and from left to right, they are from 5train and 30train respectively.}
\vspace{-1em}
\end{figure}

\begin{figure}[htbp]
\begin{minipage}[t]{0.5\linewidth}
\centering
\includegraphics[width=1.7in,height = 1.2in]{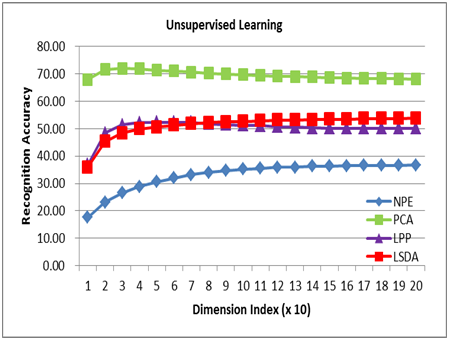}
\end{minipage}%
\begin{minipage}[t]{0.5\linewidth}
\centering
\includegraphics[width=1.7in,height = 1.2in]{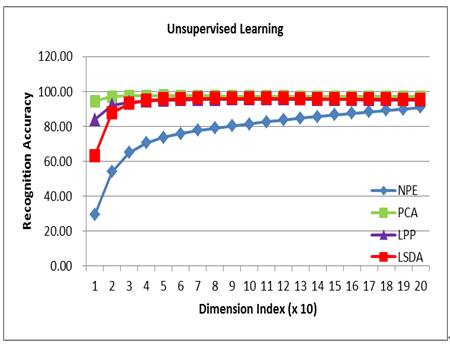}
\end{minipage}
\caption{Performance comparisons on COIL-100 database with different dimensions, and from left to right, they are from 5train and 50train respectively.}
\vspace{-1em}
\end{figure}

\subsection{Experimental Results of Supervised Learning Algorithms}
In these experiments, we constructed similarity matrix $A$ by Eq.(6), and only same class neighbor samples are kept, thus, LSDA is also considered as supervised learning dimensionality reduction method. The experimental results of USPS, Yale, ORL, YaleB, CMU-PIE and Coil-100 are given on Fig.14, Fig.15, Fig.16, Fig.17, Fig.18 and Fig.19 respectively, at the same time, we also choose the best results from them to compare to baseline method, whose results are given in Table I, Table II, Table III, Table IV, and Table V respectively. The meaning of the horizontal axis and the vertical coordinates in these figures are same to Fig.8 and Fig.9. Therefore, we will analyze the proposed approach in several different aspects. First, we will discuss the effect of the number of classes. Since the dimension of LDA is affected by the number of classes C, whose dimension is at most C-1. On USPS dataset, although there are several hundred samples of each object in training dataset, the number of classes is only ten. Thus, the dimension of LDA is at most nine. Fig.14 gives its results with the change of dimensions. From it, we can observe that the the accuracy of LDA obtains improvement with the increase of dimensions, but the maximum dimension of LDA only can be nine on this dataset. At the same time, we also can know that LPP and LSDA will not be affected by the number of classes, and the best accuracy of Baseline, LPP, LDA and LSDA is 96.9\%, 96.8\%, 91.6\% and 98.3\% respectively. Among these algorithms, LSDA can always obtain the best accuracy, and we also can observe the same conditions in other figures.

Second, the effect of the number of each object in the training dataset is discussed. In our experiments, three kinds of tasks are assessed, and their accuracies are given in Table.I, Table.II, Table.III, Table.IV and Table.V respectively. From these Tables, there are mainly three observations. (1) With the increase of the number of each object in the training dataset, the accuracies of all approaches are improved; (2) LFDA effective combines the ideas of LDA and LPP,whose performace are a litter better than that of LPP and LDA. (3) The accuracy of LSDA always is the best when different datasets are utilized and different numbers of training samples are chosen. (4) Although the neighborhood graph is also reserved in LPP, the shrunk pattern is ignored in LPP. Thus, its performance is a little better than Baseline and LDA, but it is a litter worse than LSDA. Thus, the proposed approach is effective and efficient.
\begin{figure}[htbp]
\begin{center}
\includegraphics[width=2in,height = 1.3in]{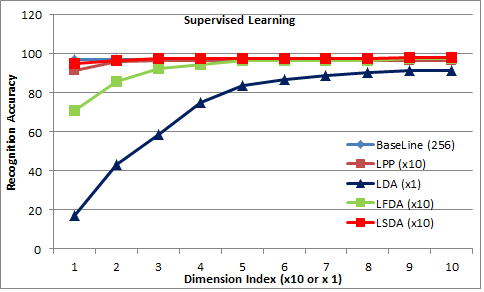}
\caption{Performance comparisons on USPS database with different dimensions, where the dimension of Baseline is 256 and the dimension of LDA changes from 1 to 9.}
\end{center}
\end{figure}

\begin{figure}[htbp]
\begin{minipage}[t]{0.5\linewidth}
\centering
\includegraphics[width=1.7in,height = 1.2in]{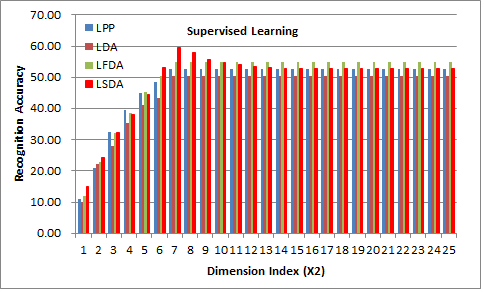}
\end{minipage}%
\begin{minipage}[t]{0.5\linewidth}
\centering
\includegraphics[width=1.7in,height = 1.2in]{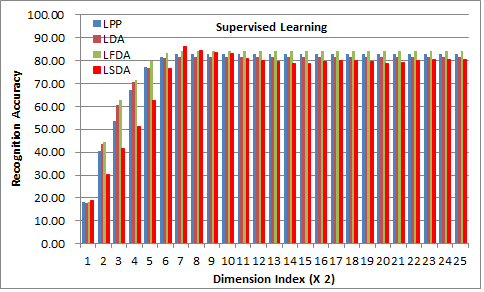}
\end{minipage}
\caption{Performance comparisons on Yale database with different dimensions, and from left to right, they are from 2train and 8train respectively.}
\vspace{-1em}
\end{figure}

\begin{figure}[htbp]
\begin{minipage}[t]{0.5\linewidth}
\centering
\includegraphics[width=1.7in,height = 1.2in]{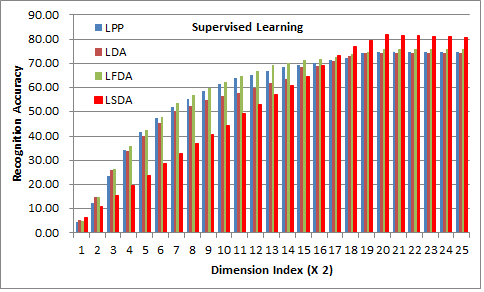}
\end{minipage}%
\begin{minipage}[t]{0.5\linewidth}
\centering
\includegraphics[width=1.7in,height = 1.2in]{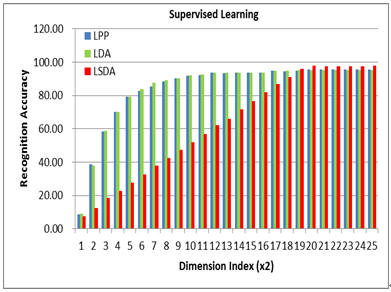}
\end{minipage}
\caption{Performance comparisons on ORL database with different dimensions, and from left to right, they are from 2train and 8train respectively.}
\vspace{-1em}
\end{figure}

\begin{figure}[htbp]
\begin{minipage}[t]{0.5\linewidth}
\centering
\includegraphics[width=1.7in,height = 1.2in]{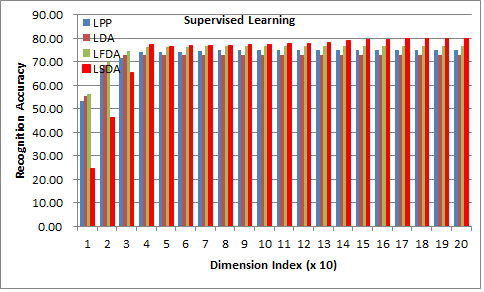}
\end{minipage}%
\begin{minipage}[t]{0.5\linewidth}
\centering
\includegraphics[width=1.7in,height = 1.2in]{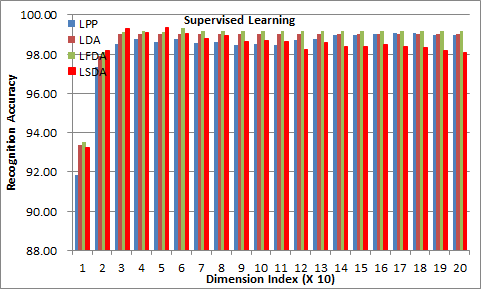}
\end{minipage}
\caption{Performance comparisons on YaleB database with different dimensions, and from left to right, they are from 2train and 8train respectively.}
\vspace{-1em}
\end{figure}

\begin{figure}[htbp]
\begin{minipage}[t]{0.5\linewidth}
\centering
\includegraphics[width=1.7in,height = 1.2in]{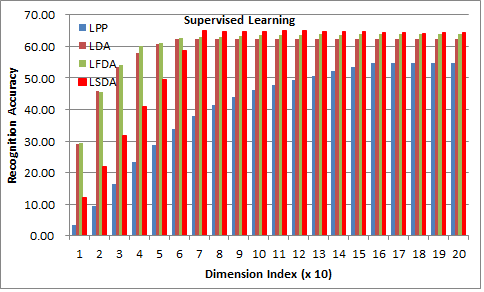}
\end{minipage}%
\begin{minipage}[t]{0.5\linewidth}
\centering
\includegraphics[width=1.7in,height = 1.2in]{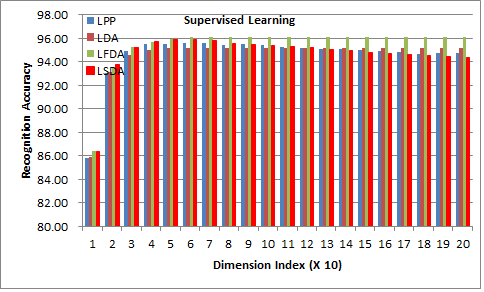}
\end{minipage}
\caption{Performance comparisons on CMU-PIE database with different dimensions, and from left to right, they are from 5train and 30train respectively.}
\vspace{-1em}
\end{figure}

\begin{figure}[htbp]
\begin{minipage}[t]{0.5\linewidth}
\centering
\includegraphics[width=1.7in,height = 1.2in]{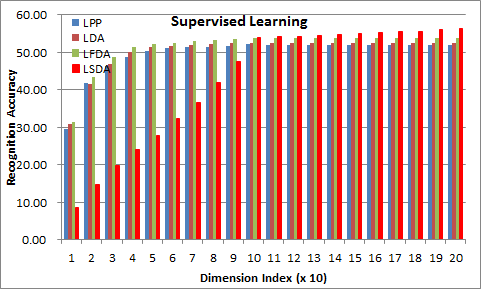}
\end{minipage}%
\begin{minipage}[t]{0.5\linewidth}
\centering
\includegraphics[width=1.7in,height = 1.2in]{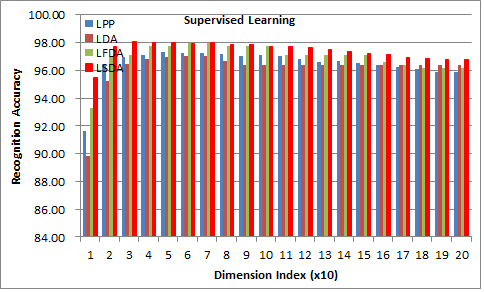}
\end{minipage}
\caption{Performance comparisons on Coil-100 database with different dimensions, and from left to right, they are from 5train and 50train respectively.}
\vspace{-1em}
\end{figure}

\subsection{The Comparison Between Unsupervised Learning Methods and Supervising Learning Method}
In above two sections, we have discussed unsupervised and supervised learning dimensionality reduction algorithms, and we can know that: 1) if the label information of samples can be employed, most of time, the performance of supervised learning algorithms is much better than that of unsupervised learning algorithms; For example, the performances of the 10th row in Table.I, Table.II, Table.III and Table.IV, which belongs to LDA, are much better than the performances of the 5-th row in these corresponding Tables, which belongs to PCA; 2) As for LPP with unsupervised and supervised learning, the difference between them is that whether the label information in the construction of similarity matrix is utilized or not, but we still can observe that the accuracy of supervised learning outperforms the accuracy of unsupervised learning, where the improvement can reach about 5\% to 30\%; 3)  Similarly, we also can find the similar case in LSDA method, where the label information of samples is very helpful, but we still observe that since the shrunk pattern is learnt in this model, which will make the data more flexible to fit the manifold structure, thus, the difference between unsupervised and supervised learning algorithms is not so great. That is to say, our proposed LSDA model is effective and stable.

\begin{table*}
\caption{ PERFORMANCE COMPARISONS ON THE YALE DATABASE WITH DIFFERENT TRAINING SAMPLES AND DIFFERENT DIMENSIONALITY REDUCTION ALGORITHMS}
\begin{center}
\begin{tabular}{|c|c|c|c|c|c|c|c|}
\hline
\ &2Train &3Train &4Train&5Train&6Train&7Train&8Train\\
\hline
\multicolumn{8}{|c|} {BaseLine (Dimension 1024)}\\
\hline
& $43.4\pm3.9$ &  $49.4\pm4.2$ & $52.6\pm4.0$& $56.2\pm4.1$&$58.7\pm4.7$&$60.2\pm4.9$&$63..6\pm5.1$\\
\hline
\multicolumn{8}{|c|} {Unsupervised Learning}\\
\hline
PCA & $42.3\pm2.4$ &  $47.9\pm5.3$ & $54.3\pm4.3$& $54.3\pm4.3$&$59.4\pm5.9$&$61.0\pm3.5$&$62.4\pm4.7$\\

NPE & $38.3\pm3.9$ &  $44.4\pm3.2$ & $54.2\pm3.7$& $54.8\pm3.5$&$60.1\pm4.1$&$58.5\pm3.0$&$59.1\pm3.5$\\

LLP & $27.7\pm1.9$ &  $31.4\pm2.8$ & $38.5\pm3.9$& $43.2\pm3.2$&$49.3\pm3.9$&$50.2\pm2.9$&$55.3\pm3.1$\\

LSDA & $46.6\pm1.8$ &  $51.8\pm2.7$ & $57.8\pm4.1$& $58.3\pm3.6$&$64.1\pm3.7$&$65.2\pm3.1$&$66.2\pm3.8$\\

\hline
\multicolumn{8}{|c|}  {Supervised Learning}\\
\hline
LPP & $52.2\pm3.5$ &  $64.3\pm3.9$ & $71.5\pm3.6$& $73.8\pm3.6$&$79.9\pm3.8$&$81.0\pm3.7$&$82.9\pm4.2$\\

LDA & $50.5\pm3.4$ &  $62.4\pm3.8$ & $69.6\pm3.7$& $72.2\pm3.2$&$77.9\pm3.8$&$80.3\pm3.5$&$81.6\pm4.0$\\

LFDA & $54.8\pm3.1$ &  $67.1\pm3.6$ & $73.9\pm3.4$& $75.5\pm3.0$&$81.6\pm3.4$&$82.9\pm3.3$&$84.1\pm4.1$\\

LSDA & $59.4\pm3.2$ &  $70.2\pm3.8$ & $77.8\pm3.5$& $78.6\pm3.0$&$84.1\pm3.6$&$86.0\pm3.3$&$86.2\pm3.7$\\
\hline
\end{tabular}
\end{center}
\end{table*}

\begin{table*}
\caption{PERFORMANCE COMPARISONS ON THE ORL DATABASE WITH DIFFERENT TRAINING SAMPLES AND DIFFERENT DIMENSIONALITY REDUCTION ALGORITHMS
}
\begin{center}
\begin{tabular}{|c|c|c|c|c|c|c|c|}
\hline
\ &2Train &3Train &4Train&5Train&6Train&7Train&8Train\\
\hline
\multicolumn{8}{|c|} {BaseLine (Dimension 1024)}\\
\hline
& $66.9\pm3.5$ &  $76.6\pm2.3$ & $82.1\pm2.2$& $86.4\pm2.4$&$88.6\pm2.3$&$91.3\pm2.4$&$92.6\pm2.4$\\
\hline
\multicolumn{8}{|c|} {Unsupervised Learning}\\
\hline
PCA & $67.2\pm3.9$ &  $76.1\pm2.0$ & $81.0\pm1.3$& $85.2\pm2.7$&$88.5\pm2.9$&$88.5\pm3.0$&$91.4\pm2.8$\\

NPE & $66.2\pm2.7$ &  $73.4\pm1.8$ & $77.9\pm2.0$& $81.4\pm2.1$&$84.1\pm2.4$&$82.8\pm2.5$&$86.5\pm2.8$\\

LLP & $55.3\pm2.4$ &  $65.3\pm2.1$ & $71.6\pm1.9$& $76.4\pm2.2$&$77.8\pm2.1$&$80.3\pm2.4$&$80.9\pm2.5$\\

LSDA & $72.2\pm2.1$ &  $79.4\pm1.3$ & $84.9\pm1.5$& $89.2\pm1.3$&$90.6\pm1.4$&$92.4\pm1.5$&$94.8\pm1.7$\\

\hline
\multicolumn{8}{|c|}  {Supervised Learning}\\
\hline
LPP & $74.7\pm3.1$ &  $84.5\pm2.4$ & $88.6\pm1.9$& $92.9\pm2.6$&$94.2\pm2.1$&$94.3\pm2.1$&$95.5\pm2.3$\\

LDA & $74.3\pm3.3$ &  $83.2\pm2.1$ & $87.5\pm1.7$& $92.1\pm2.4$&$93.4\pm2.0$&$93.8\pm1.9$&$94.9\pm2.1$\\

LFDA & $78.6\pm3.4$ &  $85.7\pm2.0$ & $90.3\pm1.6$& $93.3\pm2.2$&$94.6\pm2.1$&$94.9\pm2.0$&$96.1\pm2.3$\\

LSDA & $81.7\pm2.9$ &  $87.9\pm2.1$ & $92.1\pm1.6$& $94.6\pm2.1$&$95.9\pm1.8$&$96.4\pm1.9$&$97.6\pm2.0$\\
\hline
\end{tabular}
\end{center}
\end{table*}

\begin{table*}
\caption{PERFORMANCE COMPARISONS ON THE YALEB DATABASE WITH DIFFERENT TRAINING SAMPLES AND DIFFERENT DIMENSIONALITY REDUCTION ALGORITHMS
}
\begin{center}
\begin{tabular}{|c|c|c|c|c|c|c|}
\hline
\ &5Train &10Train &20Train&30Train&40Train&50Train\\
\hline
\multicolumn{7}{|c|} {BaseLine (Dimension 1024)}\\
\hline
& $36.4pm1.6$ &  $53.6\pm1.1$ & $69.6\pm1.1$& $77.4\pm1.2$&$81.9\pm1.0$&$84.3\pm1.5$\\
\hline
\multicolumn{7}{|c|} {Unsupervised Learning}\\
\hline
PCA & $36.6\pm1.4$ &  $52.4\pm1.3$ & $67.3\pm1.1$& $74.4\pm1.4$&$78.7\pm1.5$&$81.5\pm1.6$\\

NPE & $37.1\pm1.8$ &  $58.7\pm1.4$ & $65.6\pm1.3$& $54.8\pm1.4$&$55.8\pm1.3$&$56.7\pm1.1$\\

LLP & $42.0\pm1.6$ &  $57.1\pm1.5$ & $67.5\pm1.2$& $80.1\pm1.3$&$82.8\pm1.1$&$84.7\pm1.2$\\

LSDA & $53.3\pm0.51$ &  $88.1\pm0.42$ & $90.8\pm0.39$& $93.6\pm0.45$&$94.9\pm0.42$&$95.7\pm0.8$\\

\hline
\multicolumn{7}{|c|}  {Supervised Learning}\\
\hline
LPP & $74.9\pm1.3$ &  $88.9\pm1.0$ & $93.4\pm0.9$& $97.8\pm1.0$&$98.7\pm0.8$&$99.1\pm1.2$\\

LDA & $72.8\pm1.3$ &  $85.5\pm1.1$ & $91.5\pm0.8$& $97.1\pm0.9$&$98.1\pm0.8$&$99\pm1.1$\\

LFDA & $76.8\pm1.4$ &  $90.2\pm1.3$ & $94.8\pm1.0$& $97.9\pm0.8$&$99.1\pm0.8$&$99.4\pm1.2$\\

LSDA & $80.0\pm1.1$ &  $89.1\pm0.8$ & $95.0\pm0.8$& $97.9\pm0.7$&$98.9\pm0.6$&$99.4\pm1.0$\\
\hline
\end{tabular}
\end{center}
\end{table*}

\begin{table*}
\caption{PERFORMANCE COMPARISONS ON THE CMU-PIE DATABASE WITH DIFFERENT TRAINING SAMPLES AND DIFFERENT DIMENSIONALITY REDUCTION ALGORITHMS
}
\begin{center}
\begin{tabular}{|c|c|c|c|c|}
\hline
\ &5Train &10Train &20Train&30Train\\
\hline
\multicolumn{5}{|c|} {BaseLine (Dimension 1024)}\\
\hline
& $29.8\pm0.93$ &  $42.9\pm0.56$ & $61.4\pm0.4$& $71.5\pm1.1$\\
\hline
\multicolumn{5}{|c|} {Unsupervised Learning}\\
\hline
PCA & $28.8\pm1.1$ &  $41.8\pm0.9$ & $59.9\pm1.0$& $69.9\pm1,3$\\

NPE & $45.5\pm1.2$ &  $30.1\pm1.1$ & $85.0\pm1.2$& $90.3\pm1.0$\\

LLP & $33.1\pm0.94$ &  $41.3\pm1.2$ & $56.8\pm1.3$& $68.1\pm1.1$\\

LSDA & $48.0\pm0.71$ &  $49.0\pm0.51$ & $89.3\pm0.42$& $93.9\pm0.34$\\

\hline
\multicolumn{5}{|c|}  {Supervised Learning}\\
\hline
LPP & $54.7\pm1.0$ &  $68.0\pm0.8$ & $92.3\pm0.8$& $95.6\pm0.9$\\

LDA & $62.4\pm0.9$ &  $71.6\pm0.9$ & $93.1\pm1.0$& $95.2\pm0.9$\\

LFDA & $62.9\pm0.7$ &  $72.9\pm1.0$ & $93.7\pm0.7$& $95.7\pm0.7$\\

LSDA & $65.1\pm0.8$ &  $73.7\pm0.6$ & $94.2\pm0.5$& $95.9\pm0.7$\\
\hline
\end{tabular}
\end{center}
\end{table*}

\begin{table*}
\caption{PERFORMANCE COMPARISONS ON THE COIL-100 DATABASE WITH DIFFERENT TRAINING SAMPLES AND DIFFERENT DIMENSIONALITY REDUCTION ALGORITHMS
}
\begin{center}
\begin{tabular}{|c|c|c|c|c|c|c|}
\hline
\ &5Train &10Train &20Train&30Train&40Train&50Train\\
\hline
\multicolumn{7}{|c|} {BaseLine (Dimension 1024)}\\
\hline
& $68.0\pm0.25$ &  $78.8\pm0.39$ & $88.6\pm1.2$& $92.3\pm0.39$&$94.6\pm0.5$&$95.7\pm1.0$\\
\hline
\multicolumn{7}{|c|} {Unsupervised Learning}\\
\hline
PCA & $72.0\pm1.4$ &  $83.3\pm1.3$ & $91.6\pm1.4$& $94.9\pm1.4$&$96.6\pm1.5$&$97.6\pm1.6$\\

NPE & $36.7\pm1.8$ &  $41.9\pm1.4$ & $72.4\pm1.3$& $82.1\pm1.4$&$87.3\pm1.3$&$90.9\pm1.1$\\

LLP & $52.4\pm1.6$ &  $57.5\pm1.5$ & $86.7\pm1.2$& $91.6\pm1.3$&$94.0\pm1.1$&$95.6\pm1.2$\\

LSDA & $53.9\pm0.51$ &  $60.4\pm0.42$ & $88.0\pm0.39$& $92.9\pm0.45$&$95.4\pm0.42$&$95.9\pm0.8$\\

\hline
\multicolumn{7}{|c|}  {Supervised Learning}\\
\hline
LPP & $52.1\pm0.9$ &  $61.4\pm1.0$ & $92.5\pm1.1$& $95.1\pm1.2$&$96.6\pm1.2$&$97.3\pm1.4$\\

LDA & $52.4\pm0.9$ &  $60.4\pm1.1$ & $91.8\pm1.0$& $94.8\pm1.3$&$96.3\pm1.2$&$97.1\pm1.1$\\

LFDA & $54.5\pm1.0$ &  $62.8\pm1.0$ & $92.5\pm1.1$& $95.8\pm1.1$&$96.3\pm1.2$&$97.8\pm1.2$\\

LSDA & $56.3\pm0.7$ &  $64.7\pm0.9$ & $92.7\pm0.9$& $95.8\pm1.0$&$97.3\pm1.1$&$98.1\pm1.1$\\
\hline
\end{tabular}
\end{center}
\vspace{-2em}
\end{table*}

\subsection{Parameter Sensitivity Analysis}
In above Sections, we have proved the effective of LSDA, but in this Section, we will further evaluate the parameter sensitivity of it. Due to space limitation, we only choose one dataset for each task, and their performances on USPS, YaleB and Coil-100 datasets are shown in Fig.20, Fig.21 and Fig.22, Fig.23 and Fig.24  respectively.  In these figures, Fig.20 (a), Fig.21 (a),Fig.22 (a) and Fig.23 (a) are the performance variance of sigma when gamma and dimension are fixed, where all gammas are equal to $2^{-10}$, Fig.20 (b), Fig.21(b), Fig.22(b) and Fig.23 (b) are the performance variance of  gamma when sigma and dimension are fixed, where all sigmas are set to 0.9, 0.3 and 0.9 respectively. Further, for USPS and Coil-100, the dimension is set to 100, but the dimension of YaleB is assigned to 200.

There are mainly two observations from these figures. 1) When Gamma is fixed, the performance of LSDA is stable with the change of Sigma, especial for intermediate value; 2) When Sigma is fixed, the performance of LSDA has some fluctuation with the change of Gamma, and both ends also have the best performance, 3) The number of each subject sample in the training dataset will slightly affect the choice of Gamma. Thus, the optimization of LSDA will be easier and we can quickly choose the best parameters.

\begin{figure}[htbp]
\begin{minipage}[t]{0.5\linewidth}
\centering
\includegraphics[width=1.7in,height = 1.2in]{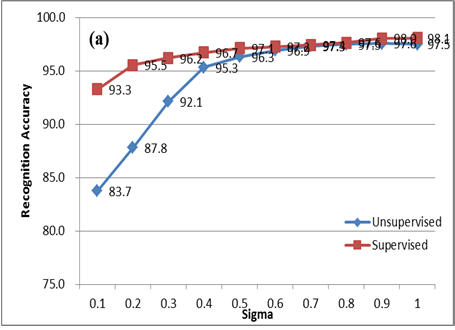}
\end{minipage}%
\begin{minipage}[t]{0.5\linewidth}
\centering
\includegraphics[width=1.7in,height = 1.2in]{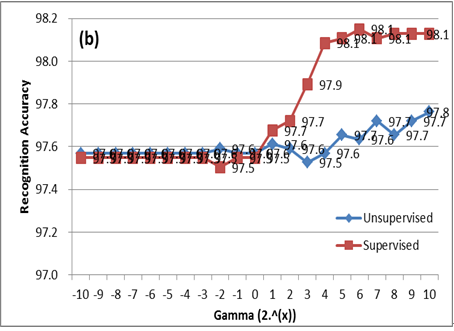}
\end{minipage}
\caption{Performance analysis on the USPS database, (a) the performance of the variation of parameter sigma when gamma and dimension are fixed to $2^{-10}$ and 100 respectively, (b) the performance of the variation of parameter gamma when sigma and dimension are fixed to 0.9 and 100 respectively.}

\end{figure}

\vspace{-1em}
\begin{figure}[htbp]
\begin{minipage}[t]{0.5\linewidth}
\centering
\includegraphics[width=1.7in,height = 1.2in]{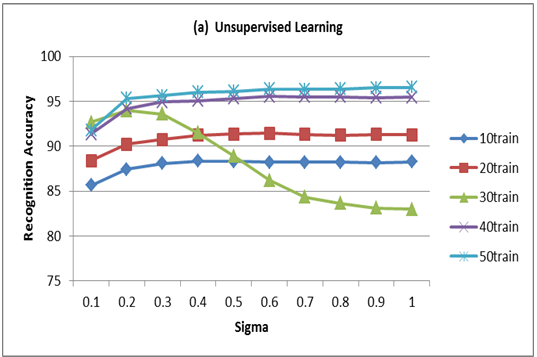}
\end{minipage}%
\begin{minipage}[t]{0.5\linewidth}
\centering
\includegraphics[width=1.7in,height = 1.2in]{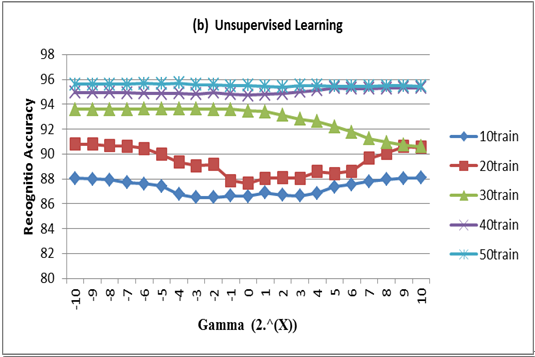}
\end{minipage}
\caption{ Performance analysis of unsupervised learning on the YaleB database, the performance of the variation of parameter sigma when gamma and dimension are fixed to $2^{-10}$ and 200 respectively.}
\vspace{-2em}
\end{figure}

\begin{figure}[htbp]
\begin{minipage}[t]{0.5\linewidth}
\centering
\includegraphics[width=1.7in,height = 1.3in]{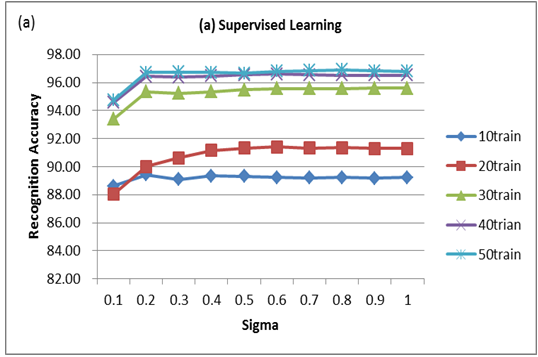}
\end{minipage}%
\begin{minipage}[t]{0.5\linewidth}
\centering
\includegraphics[width=1.7in,height = 1.3in]{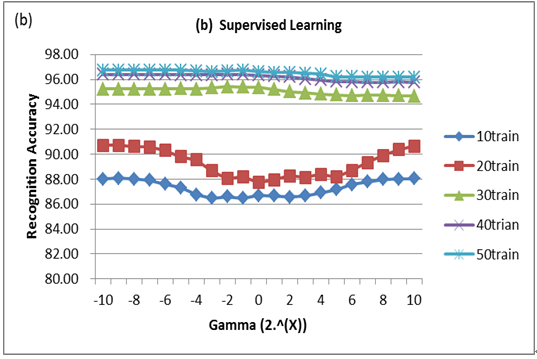}
\end{minipage}
\caption{ Performance analysis of supervised learning on the YaleB database,the performance of the variation of parameter gamma when sigma and dimension are fixed to 0.3 and 200 respectively.}

\end{figure}

\begin{figure}[htbp]
\begin{minipage}[t]{0.5\linewidth}
\centering
\includegraphics[width=1.7in,height = 1.2in]{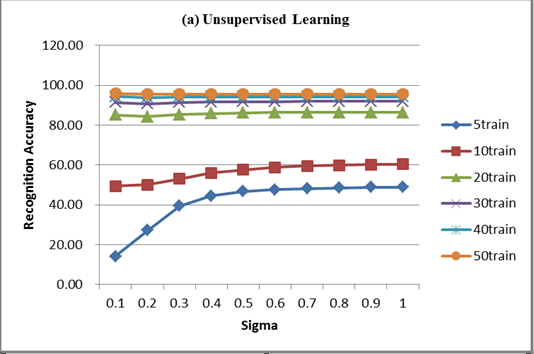}
\end{minipage}%
\begin{minipage}[t]{0.5\linewidth}
\centering
\includegraphics[width=1.7in,height = 1.2in]{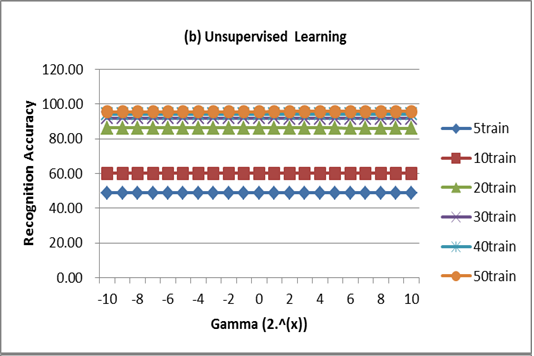}
\end{minipage}
\caption{ Performance analysis of unsupervised learning on the Coil-100 database, the performance of the variation of parameter sigma when gamma and dimension are fixed to $2^{-10}$ and 100 respectively.}

\end{figure}

\begin{figure}[htbp]
\begin{minipage}[t]{0.5\linewidth}
\centering
\includegraphics[width=1.7in,height = 1.2in]{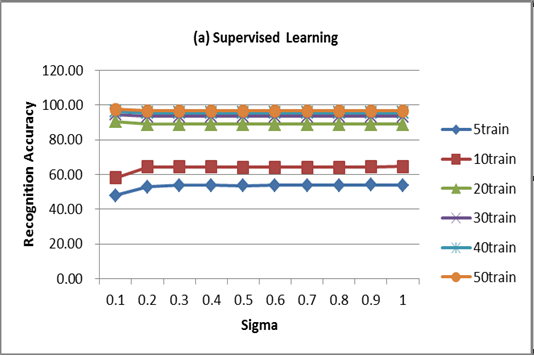}
\end{minipage}%
\begin{minipage}[t]{0.5\linewidth}
\centering
\includegraphics[width=1.7in,height = 1.2in]{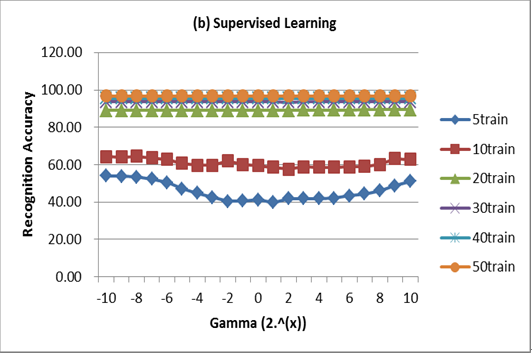}
\end{minipage}
\caption{ Performance analysis of supervised learning on the Coil-100 database, the performance of the variation of parameter gamma when sigma and dimension are fixed to 0.9 and 100 respectively.}
\end{figure}

\subsection{The Generalization Analysis}
From Subection C in algorithm analysis of Section III, we can know that the objective function LSDA has strong generalized form, and LDA and local LDA are the special case of it when different extreme parameters are utilzied. Thus, in this experiment, we will assess their performances on USPS, YaleB and Coil-100 datasets when extreme gamma parameters are utilized, and the results are shown in Fig.25, Fig.26 and Fig.27 respectively. In the optimization, and the dimensions on all datasets are set to 200, and sigma is set to 0.7. From them, we can observe that when gamma is zero or positive infinity respectively, their accuracies are still comparable to the performance of LDA or PCA, but the optimization parameter of LSDA can obtain the best performance.For example, the accuracies of LSDA, LSDA (Gamma=0) and LSDA (Gamma = INF) are 97.8\%, 94.6\% and 95.4\% respectively in Fig.25 where the dimension is equal to 60, and we also can observe the same case in other figures. Thus, it further proves that our proposed approach has better generalized performance.

\begin{figure}[htbp]
\begin{minipage}[t]{0.5\linewidth}
\centering
\includegraphics[width=1.7in,height = 1.2in]{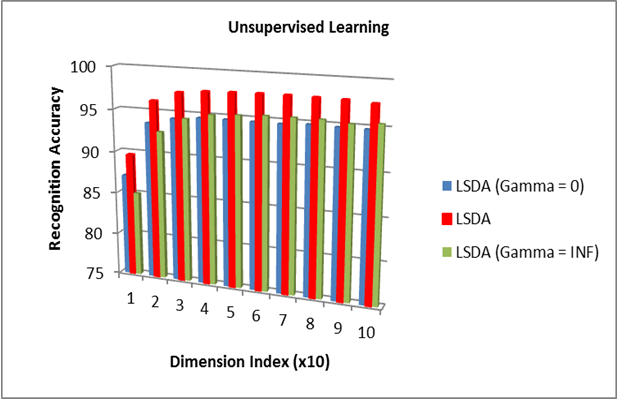}
\end{minipage}%
\begin{minipage}[t]{0.5\linewidth}
\centering
\includegraphics[width=1.7in,height = 1.2in]{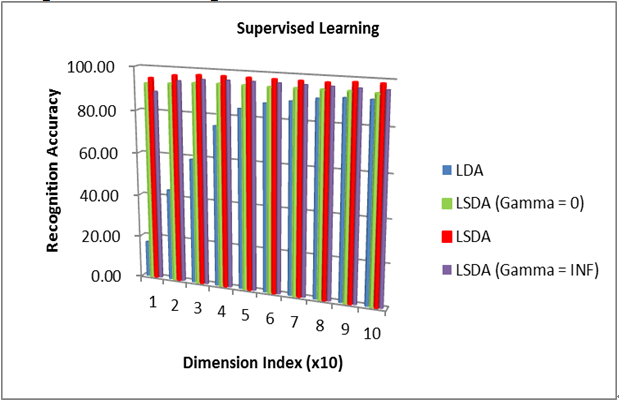}
\end{minipage}

\caption{ Performance analysis on USPS database when Gamm=0, Gamma = INF and the optimized Gamma are employed respectively, and from left to right, they are unsupervised and supervised learning algorithms respectively.}
\end{figure}

\begin{figure}[htbp]
\begin{minipage}[t]{0.5\linewidth}
\centering
\includegraphics[width=1.7in,height = 1.2in]{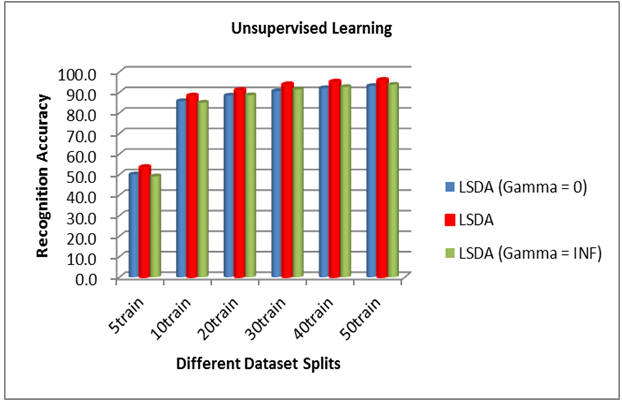}
\end{minipage}%
\begin{minipage}[t]{0.5\linewidth}
\centering
\includegraphics[width=1.7in,height = 1.2in]{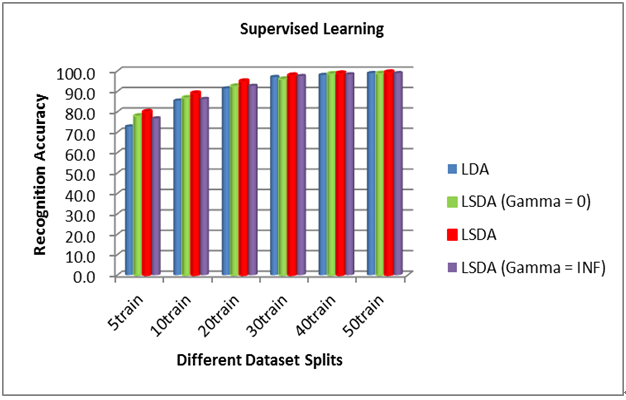}
\end{minipage}
\caption{ Performance analysis on YaleB database when Gamm=0, Gamma = INF and the optimized Gamma are employed respectively, and from left to right, they are unsupervised and supervised learning algorithms respectively.}
\end{figure}

\begin{figure}[htbp]
\begin{minipage}[t]{0.5\linewidth}
\centering
\includegraphics[width=1.7in,height = 1.2in]{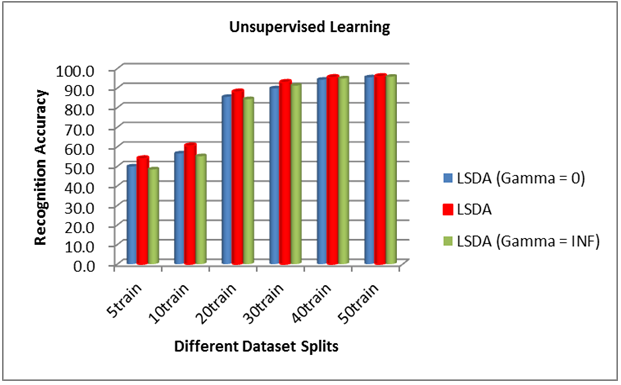}
\end{minipage}%
\begin{minipage}[t]{0.5\linewidth}
\centering
\includegraphics[width=1.7in,height = 1.2in]{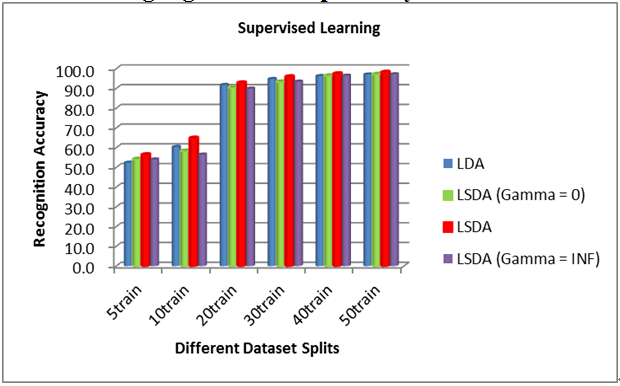}
\end{minipage}
\caption{ Performance analysis on Coil-100 database when Gamm=0, Gamma = INF and the optimized Gamma are employed respectively, and from left to right, they are unsupervised and supervised learning algorithms respectively.}
\vspace{-1em}
\end{figure}

\section{CONCLUSION}
In this paper, we have proposed a novel and universal unsupervised and supervised learning dimensionality reduction method. It is mainly based on pattern shrinking technique. The main idea is to simultaneously learn the pattern shrinking and the projector matrix, and make the data more flexible to fit the manifold structure, which is more convenient for non-Gaussian distribution data and real-life data. The advantage of our method is three-fold. First, uncovering the manifold structure can be mined by our method. Particularly, the shrunk pattern learned by the proposed algorithm does not have the orthogonal constraint, which makes it more flexible to fit the manifold structure. The learned manifold knowledge is particularly helpful for achieving better dimensionality reduction result. Second, the transformation matrix is learnt from the original space and the pattern shrinking space, which contributes to more precise structural information for dimensionality reduction and recognition. Third, the pattern shrinking and transformation matrix are simultaneously learnt, which makes it easy to reduce dimension and data representation. Experimental results on several datasets show that when compared with the state-of-the-art unsupervised and supervised learning dimensionality reduction methods, it performs better. Moreover, it has much better generalized form, and LDA and local LDA are the special case of it. In addition, an efficient optimization algorithm is introduced to solve the non-convex objective function with low computational cost. Thus, LSDA is effective and efficient.

\ifCLASSOPTIONcaptionsoff
  \newpage
\fi



%

\begin{IEEEbiography}[{\includegraphics[width=1in,height=1.25in,clip,keepaspectratio]{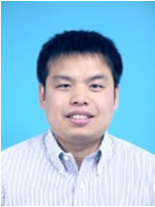}}]{Zan Gao}
is an associate professor in the school of Computer and Communication engineering, key laboratory of computer vision and system, Ministry of Education, Tianjin University of Technology. From Sep. 2009 to Sep. 2010, he was a visiting scholar in the School of Computer Science, Carnegie Mellon University, USA. He received his Ph.D degree from Beijin University of Posts and Telecommunications in 2011. His research interests include computer vision, multimedia analysis and retrieval.
\end{IEEEbiography}

\begin{IEEEbiography}[{\includegraphics[width=1in,height=1.25in,clip,keepaspectratio]{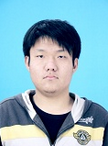}}]{Guo-tai Zhang} is pursuing his master degree in the school of Computer and Communication engineering, Tianjin University of Technology. He received his Bachelor degree from Shandong Sports University in 2013. His research interests include computer vision, multimedia analysis and retrieval.
\end{IEEEbiography}

\begin{IEEEbiography}[{\includegraphics[width=1in,height=1.25in,clip,keepaspectratio]{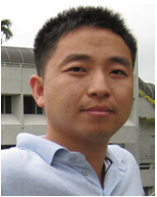}}]{Fei-ping NIE}
received the Ph.D. degree in Computer Science from Tsinghua University, China in 2009. His research interests are machine learning and its application fields, such as pattern recognition, data mining, computer vision, image processing and information retrieval. He has published more than 100 papers in the following top journals and conferences: TPAMI, IJCV, TIP, TNNLS/TNN, TKDE, TKDD, TVCG, TCSVT, TMM, TSMCB/TC, Machine Learning, Pattern Recognition, Medical Image Analysis, Bioinformatics, ICML, NIPS, KDD, IJCAI, AAAI, ICCV, CVPR, SIGIR, ACM MM, ICDE, ECML/PKDD, ICDM, MICCAI, IPMI, RECOMB. According to the Google scholar, his papers have been cited more than 2000 times. He is now serving as Associate Editor or PC member for several prestigious journals and conferences in the related fields.
\end{IEEEbiography}

\begin{IEEEbiography}[{\includegraphics[width=1in,height=1.25in,clip,keepaspectratio]{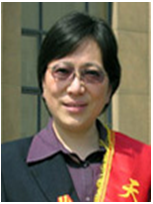}}]{Hua Zhang}
is a professor in the school of Computer and Communication Engineering, Tianjin University of Technology, Tianjin, China. She received her doctor degree from Tianjin University in 2008. Her research interests include multimedia analysis and virtual reality
\end{IEEEbiography}

%




\end{document}